\pdfoutput=1
\documentclass{article}
\usepackage{ijcai21}

\usepackage{times}
\usepackage{soul}
\usepackage[hidelinks]{hyperref}
\usepackage[utf8]{inputenc}
\usepackage{subfigure}
\usepackage{url}
\usepackage[hidelinks]{hyperref}
\usepackage[utf8]{inputenc}
\usepackage[small]{caption}
\usepackage{graphicx}
\usepackage{amsmath}
\usepackage{amsthm}
\usepackage{booktabs}
\usepackage{tabularx}
\usepackage{multirow}
\usepackage{helvet}
\usepackage{algorithm}
\usepackage{algorithmic}
\usepackage{color}
\usepackage{amsfonts,amsmath,amssymb,amsthm,mathtools}
\usepackage{wrapfig}
\usepackage{courier}
\usepackage[switch]{lineno}

\usepackage{lipsum}

\title{MapGo: Model-Assisted Policy Optimization for Goal-Oriented Tasks}

\author{
Menghui Zhu$^{1\,2}\,$\footnote{Equal contribution. This work was done when the first author was an intern at Tencent. \dag ~marks the corresponding author. Our code is available at \url{https://github.com/apexrl/MapGo}.}
\and
Minghuan Liu$^{1\,*}$ \and
Jian Shen$^{1}$ \and
Zhicheng Zhang$^{1}$ \and
Sheng Chen$^{2}$  \and \\
Weinan Zhang$^{1,\dag}$ \and
Deheng Ye$^{2}$ \and
Yong Yu$^{1}$ \and
Qiang Fu$^{2}$ \And
Wei Yang$^{2}$
\affiliations
$^1$Shanghai Jiao Tong University, Shanghai, China\\
$^2$Tencent AI Lab, Shenzhen, China\\
\emails
\{zerozmi7,minghuanliu,wnzhang\}@sjtu.edu.cn,
\{victchen,dericye,leonfu,willyang\}@tencent.com
}

\newcommand{\citea}[1]{\citeauthor{#1} \shortcite{#1}}

\usepackage{amsmath,amsfonts,bm}

\def\eqref#1{equation~\ref{#1}}

\def\1{\bm{1}}

\DeclareMathAlphabet{\mathsfit}{\encodingdefault}{\sfdefault}{m}{sl}
\SetMathAlphabet{\mathsfit}{bold}{\encodingdefault}{\sfdefault}{bx}{n}

\newcommand{\fig}[1]{Fig.~\ref{#1}}
\newcommand{\eq}[1]{Eq.~(\ref{#1})}

\newcommand{\alg}[1]{Algo.~\ref{#1}}

\newcommand{\citet}[1]{\citeauthor{#1}~\shortcite{#1}}

\newcommand{\bbE}{\ensuremath{\mathbb{E}}} 
\newcommand{\bbR}{\ensuremath{\mathbb{R}}} 
\newcommand{\caD}{\ensuremath{\mathcal{D}}}

\usepackage{xcolor}

\begin{document}

\maketitle

\begin{abstract}
In Goal-oriented Reinforcement learning, relabeling the raw goals in past experience to provide agents with hindsight ability is a major solution to the reward sparsity problem. 
In this paper, 
to enhance the diversity of relabeled goals, we develop FGI (Foresight Goal Inference), a new relabeling strategy that relabels the goals by looking into the future with a learned dynamics model. 
Besides, to improve sample efficiency, we propose to use the dynamics model to generate simulated trajectories for policy training. 
By integrating these two improvements, we introduce the MapGo framework (Model-Assisted Policy Optimization for Goal-oriented tasks). 
In our experiments, we first show the effectiveness of the FGI strategy compared with the hindsight one, and then show that the MapGo framework achieves higher sample efficiency when compared to model-free baselines on a set of complicated tasks.
\end{abstract}

\section{Introduction}

In this paper, we consider the problem of Goal-oriented Reinforcement Learning (GoRL), where agents are expected to complete a series of similar tasks and learn to reach multiple goals. 
Typically, the reward is designed as binaries to indicate whether the agent reaches the goal, which, however, is too sparse for the agent to learn a good policy efficiently. 
To that end, in the literature, several works have been proposed to relabel or re-generalize different goals along past trajectories~\cite{andrychowicz2017hindsight,fang2018dher,fang2019curriculum}, trying to reuse the past failed experiences by relabeling them with reachable goals. 
An early-stage example, Hindsight Experience Replay (HER)~\cite{andrychowicz2017hindsight}, is inspired by human's hindsight ability that one can always learn from an undesired outcome as from a desired one. 
Besides, other works \cite{florensa2018automatic,ren2019exploration} pay attention to choosing suitable behavioral (training-time) goals to help explore the environment in a more efficient manner.

However, existing works have two significant issues. 
First, most of them are model-free methods that are considered less sample-efficient at most times.
In contrast, model-based reinforcement learning (MBRL) has shown better sample efficiency compared to the model-free counterparts \cite{langlois2019benchmarking}. 
This motivates us to investigate the feasibility of model-based techniques to GoRL. 
The second issue lies in the limited guidance ability of hindsight goals \cite{andrychowicz2017hindsight}. Hindsight goals are usually sampled from trajectories of a past policy, which may not be appropriate for the current policy.
Also, the diversity of the selected goals is limited since the relabeled goals and the updated tuples $(s,a,s',g)$ are chosen from the same trajectory.
Therefore, generating more diverse goals is expected.

Based on these considerations, we develop a model-based framework for GoRL, called MapGo (Model-Assisted Policy optimization for Goal-Oriented tasks). In MapGo, we learn a bootstrapped ensemble of dynamics models \cite{chua2018deep} to help tackle the two issues above. 
First, we propose the Foresight Goal Inference (FGI) strategy, which uses the dynamics models to relabel the goals by looking into the future. Besides, the learned dynamics models are further used to generate branched short rollouts for policy optimization in a Dyna-style manner \cite{janner2019trust}. 

We conduct extensive experiments on a set of continuous control tasks. Specifically, we first compare the relabeled goals of FGI with HER \cite{andrychowicz2017hindsight} to verify the efficacy of FGI. Then we evaluate the MapGo framework on complicated goal-oriented learning benchmarks, indicating the higher sample efficiency compared with former model-free algorithms. Finally, we make a comprehensive ablation study to analyze the performance improvement of MapGo while also examining its limitation. 

In a nutshell, our contributions are: 

(i) We design a novel goal relabeling mechanism called FGI that provides foresight goals with diversity. 

(ii) Based on FGI, we propose the MapGo framework, which incorporates model-based training techniques into GoRL to improve the sample efficiency.

(iii) We provide an in-depth understanding of the components of MapGo for contributing to the final performance.

\section{Background}

\subsection{Goal-oriented Reinforcement Learning}
Compared with standard Reinforcement Learning (RL) that is usually modeled as Markov Decision Process (MDP), Goal-oriented Reinforcement Learning (GoRL) augments it with a goal state. Formally, the Goal-oriented MDP (GoMDP) is denoted as a tuple $\langle \mathcal{S}, \mathcal{A}, M^*, \mathcal{G}, r_g, \gamma, \phi, \rho_0, p_g \rangle$, where $\mathcal{S}\in\bbR^{d_s}$ and $\mathcal{A}\in\bbR^{d_a}$ are the state space and the action space respectively, $M^*: \mathcal{S} \times \mathcal{A} \times \mathcal{S} \rightarrow [0, 1]$ represents the transition function, $\gamma \in [0,1)$ is the discount factor, $\mathcal{G}\in\bbR^{d_g}$ denotes the goal space and $\phi: \mathcal{S} \rightarrow \mathcal{G}$ is a tractable mapping function that provides goal representation. $\rho_0$ is the initial state distribution and $p_g$ is the distribution of desired goals. In GoRL, a \textit{desired goal} is a task to solve which are normally given by the environments; and a \textit{behavioral goal} indicates the target used for sampling. Without carefully designed goal selection mechanism, these two goals are generally the same.

In GoMDP, the reward function $r:\mathcal{S} \times \mathcal{A} \times \mathcal{G} \rightarrow \mathbb{R}$ is goal-conditioned determining whether the goal is reached:
\begin{equation}\label{eq:goal-reward}
    r_g(s_t,a_t, g) = 
    \left\{
    \begin{aligned}
     0, &\quad \|\phi(s_{t+1})-g\| \leq \epsilon \\  
    -1, &\quad\text{ otherwise}    
    \end{aligned}
    \right. ~.
\end{equation}
Therefore, agents can utilize a universal value function \cite{schaul2015universal} that uses a single function approximator to represent a set of goal-based value functions. Given a state $s_t\in\mathcal{S}$ and a goal $g\in\mathcal{G}$, the value under the policy $\pi$ is defined as: 
\begin{equation}
    V^{\pi}(s_t, g) = \mathbb{E}_{\substack{a_t \sim \pi\\s_{t+1} \sim M^*}} \Big[ \sum_{t}^{\infty} \gamma^t r(s_t,a_t,g) \Big]~.
\end{equation}
The objective of GoRL is to find a goal-based policy $\pi(a|s,g)$ to maximize the expectation of discounted cumulative rewards over the distribution of $p_g$: 
\begin{equation}
    \eta = {\mathbb{E}}_{\substack{a_t \sim \pi\\s_{t+1} \sim M^*\\ g\sim p_g}} \Big[\sum_{t=0}^{\infty} \gamma^t r(s_t,a_t,g) \Big]~.
\end{equation}

\subsection{Hindsight Relabeling}

In GoRL, agents are provided with a specific goal from environments at the beginning of an episode. Specifically, the agent first receives a behavioral goal $g$ (from the environment or self-generated) at an initial state $s_0$, then it starts to interact with the environment for $T$ timesteps and generates a trajectory $\{(s_0, a_0, s_1, g, r_1), ..., (s_T, a_T, s_{T+1}, g, r_{T+1})\}$. Due to the sparse reward signal shown in \eq{eq:goal-reward}, the agent will not get any valuable reward $r_t$ until it reaches the goal state. Therefore, \citet{andrychowicz2017hindsight} proposed Hindsight Experience Replay (HER) to relabel the goals for those failed trajectories so that the agent can learn from failures. The often-used $future$ strategy randomly chooses a state $s_{t+k}$ after the relabeled state $s_t$; the $final$ strategy of HER is relabeling with the final state $s_T$ of the corresponding trajectory; and the $episode$ strategy instead chooses several states in the whole corresponding trajectory. Unlike these prior works, we may not try to find reasonable goals from historical data but let the agent predict and relabel the goals that it can achieve by building an environment dynamics model.

\section{Methodology}

In this section, we present our model-assisted policy optimization framework and explain how foresight goal inference works along with universal model-based policy optimization.

\begin{figure}[!t]
    \centering
    \includegraphics[width=0.8\linewidth]{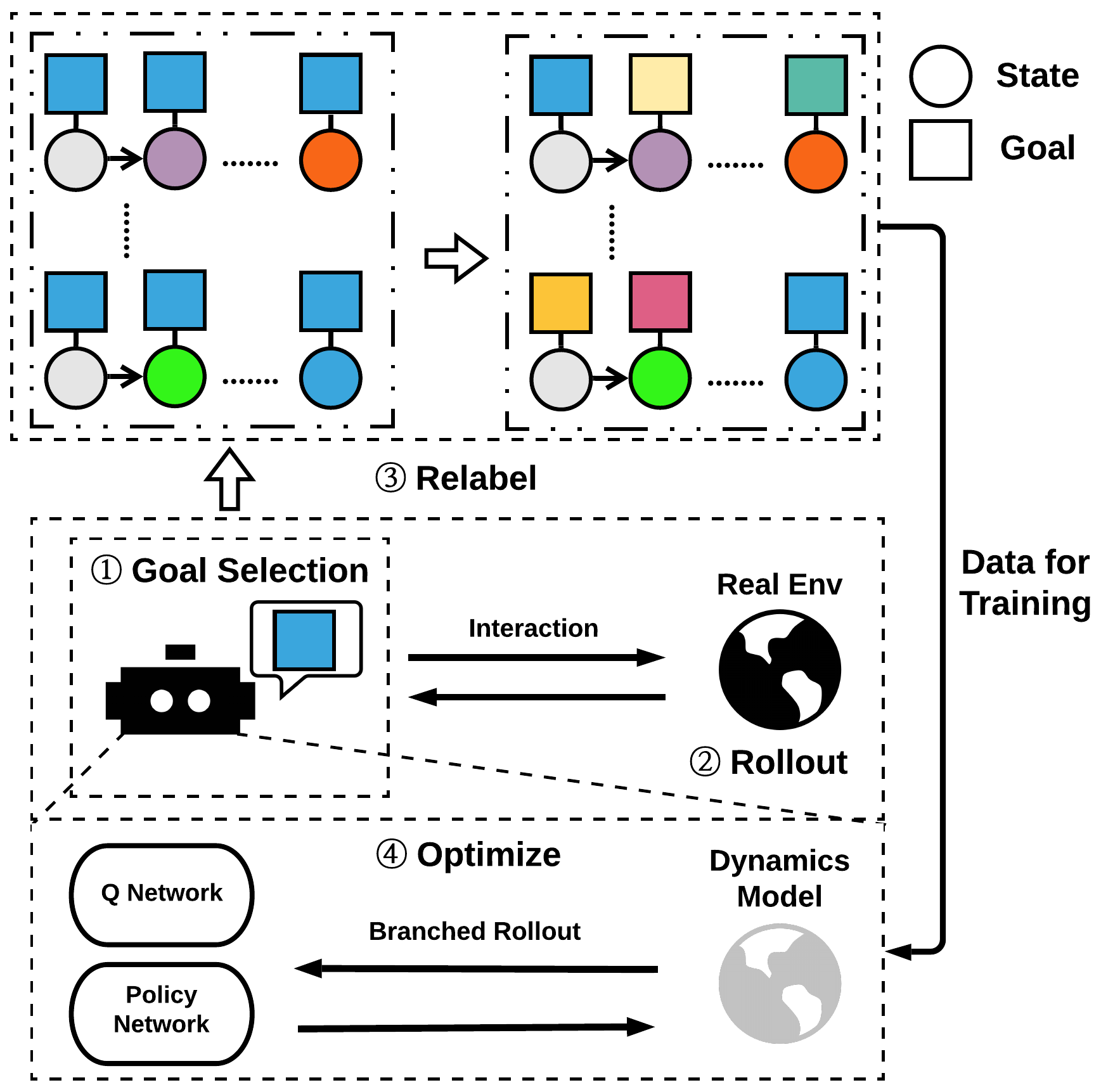}
    \caption{The overall framework of MapGo, containing four stages: i) \textit{goal selection}: the agent selects a behavioral goal for sampling; ii) \textit{rollout}: the agent performs rollout in the environment to generate trajectories;  iii) \textit{relabeling}: before training, trajectories samples from the environment are relabeled iv) \textit{optimization}: the agent optimizes the policy and value networks. Different colors represent different states. Compared with other GoRL algorithms, MapGo focuses on enhancing the \textit{relabeling} and \textit{optimization} stages.}
    \label{fig:mbgl}
\end{figure}

\subsection{Overall Framework}
We illustrate the overall framework of our Model-Assisted Policy Optimization for Goal-Oriented Tasks (MapGo) in \fig{fig:mbgl}. 
The agent starts with selecting the behavioral goals at an initial state and interacts with the environment to collect rollouts conditioned on the goal. Such experiences will then be relabeled by informative goal states, and the rewards will also be recomputed based on the relabeled goals, which will further be utilized for optimizing the policy and the value functions of the agent using off-policy RL algorithms. Our MapGo framework contributes mainly to the \textit{relabeling} and \textit{optimization} stage.

MapGo involves two functional components: i) Foresight Goal Inference (FGI), which is used to infer an appropriate goal for the agent in the \textit{relabeling} stage; 
and ii) Universal Model-based Policy Optimization (UMPO), which is used to train the goal-based policy via generated short rollouts together with the experience sampled from the environment in the \textit{optimization} stage. 
We detail the whole algorithm in \alg{alg:MBGL}, where \texttt{FGI} and \texttt{UMPO} represent the above two modules respectively. Our framework leaves flexible choices for \textit{goal selection} stage and we denote it as \texttt{GoalGen}. 

Below we move on to the details of \texttt{FGI} and \texttt{UMPO}.

\begin{algorithm}[t]
\caption{Model-Assisted Policy Optimization (MapGo)}
\label{alg:MBGL}
\begin{algorithmic}
\REQUIRE Policy parameter $\theta$, Q value parameter $\omega$, and dynamics model parameter $\psi$
\STATE $\caD_{\text{env}} \leftarrow \varnothing$; 
        \FOR{$i=1$ to $K$}
            \STATE $(s_0,g) \leftarrow \texttt{GoalGen}(\caD_{\text{env}}, \mathcal{T}^* , \pi_{\theta});$
            \FOR{$t=0$ to $h-1$}
                \STATE $a_t \leftarrow \pi_{\theta}(s_t,g);$
                \STATE $s_{t+1} \sim M^*(\cdot|s_t,a_t);$
                \STATE $r_t \leftarrow r_g(s_t,a_t,s_{t+1});$
            \ENDFOR
            \STATE $\tau \leftarrow \{s_0,a_0,r_0,s_1,a_1,r_1...\};$
            \STATE $\caD_{\text{env}} \leftarrow \caD_{\text{env}}\cup \{\tau\};$
            \STATE Update $M_{\psi}$ according to \eq{dynamic_model_loss};
        \STATE $\caD_{\text{real}} \leftarrow \texttt{FGI}(M_{\psi}, \caD_{\text{env}}, \pi, \phi, r)$;
        \STATE $\pi_{\theta}, Q_{\omega} \leftarrow \texttt{UMPO}(\pi_{\theta}, Q_{\omega}, \caD_{\text{env}}, \caD_{\text{real}}, M_{\psi})$;
        \ENDFOR
\end{algorithmic}
\end{algorithm}

\subsection{Foresight Goal Inference}
Foresight Goal Inference (FGI) serves for the \textit{relabeling} stage, which is designed for providing future reached goals w.r.t. the current policy based on the historical experience.

\subsubsection{Dynamics Modeling}
We implement such a foresight ability with simulated rollouts by utilizing an environment dynamics model. 
Generally, the environment dynamics, i.e., transition function $M^*$ is assumed to be unknown in RL. 
However, we can build an approximated dynamics model $M$ during interactions with the environment by executing a policy $\pi(a|s)$:
\begin{equation}
    M(s'|s,a) = \bbE_{(s,a,s')\sim\caD}[P(s'|s,a)]~,
\end{equation}
where $\caD$ is a replay buffer in which the experiences are sampled by the policy $\pi$. Specifically, we use a bootstrapped probabilistic dynamics model~\cite{chua2018deep}, which can capture both aleatoric uncertainty and epistemic uncertainty.

Formally, we denote $M_{\psi}$ as a parameterized dynamics model and $\{M_{\psi}^i\}_{N}$ as its ensemble. For each $M_{\psi}^i$, it approximates the transition probability $p(s' |s,a)$ as a Gaussian distribution with diagonal covariances $\mathcal{N}(\mu_{\psi}^{i}(s, a), \Sigma_{\psi}^{i}(s, a))$ given a state $s$ and an action $a$. The models are trained by minimizing the negative log likelihood:
\begin{equation}
\begin{aligned}
    \label{dynamic_model_loss}
    \mathcal{L}(\psi) = \sum_{n=1}^{N}\left[\mu_{\psi}\left(s_{n}, a_{n}\right)-s_{n+1}\right]^{\top} \Sigma_{\psi}^{-1}\left(s_{n}, a_{n}\right) \\
    \left[\mu_{\psi}\left(s_{n}, a_{n}\right)-s_{n+1}\right]+\log \operatorname{det} \Sigma_{\psi}\left(s_{n}, a_{n}\right)~.
\end{aligned}
\end{equation}

\begin{algorithm}[t]
\caption{Foresight Goal Inference (FGI)}
\label{alg:FGI}
\begin{algorithmic}
\REQUIRE dynamics model $f_\theta$, Dataset $\caD_{\text{env}}$, Current policy $\pi$, Mapping function $\phi$, Reward function $r$, Rollout maximum length $H$
\ENSURE Relabeled dataset $\caD_{\text{env}}'$
\FOR{every tuple $(s_t^i,a_t^i,r_t^i,g_t^i) \in \caD_{\text{env}}$}
    \STATE Let $L_i$ be the length of the trajectory of the tuple;
    \STATE Random choose a rollout step $h$ uniformly in $(1, L_i)$;
    \IF{$L_i>H$}
    \STATE Relabeling $g_t^i$ with $\phi(s^i_{t+H})$ by \textit{future} strategy HER;
    \ELSE
    \STATE Choose a new behavioral goal $g^i_{new}$ by \textit{future} strategy HER;
    \STATE Adopt $h$ steps rollouts on $s^i_{t+1}$ using $\hat a^i_{t'}\sim\pi(\cdot|\hat s^i_{t'}, g^i_{new})$ in $M_\psi$ , $ t+1 \leq t' \leq t+h$ and get $\hat s^i_{t+h+1}$;
    \STATE Relabeling $g^i_t$ with $\phi(\hat s^i_{t+h+1})$;
    \ENDIF
    \STATE Calculate the new reward $r'_i=r_g(s^i_t,a^i_t,g^i_t)$;
\ENDFOR

\end{algorithmic}
\end{algorithm}

\subsubsection{Goal Inference}

\begin{figure}[t!]
    \centering
    \includegraphics[width=0.8\linewidth]{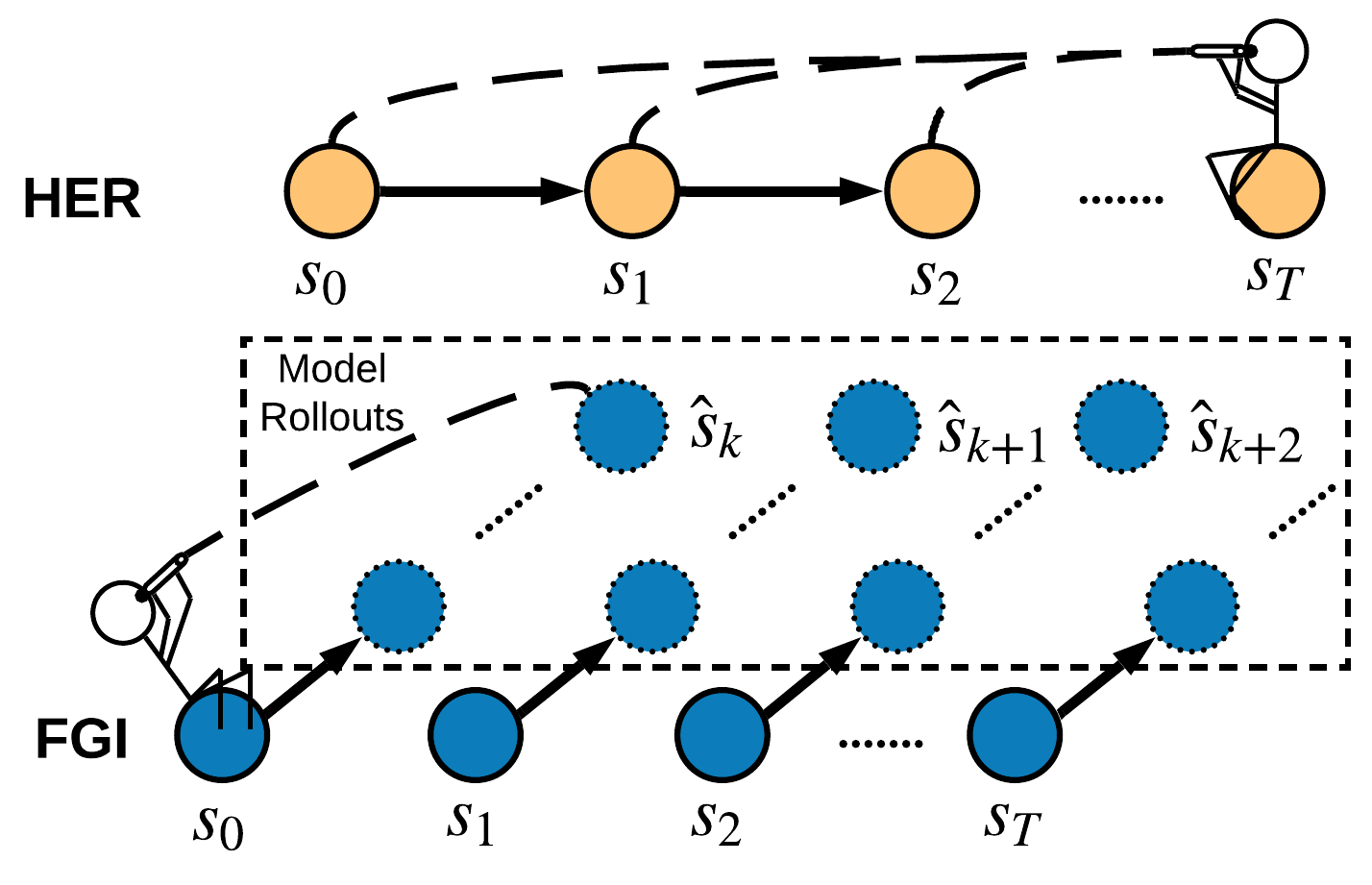}
    \caption{Illustration of the FGI and the HER relabeling strategies.}
    \label{fig:fgi}
\end{figure}

With a well-trained dynamics model, the agent is capable of foreseeing a considerable future goal. Particularly, under certain state $s_t$, we replace the original desired goal state $g_t$ with a foresight goal $\hat{g}_t$ by reasoning forward $K$ steps using an ensemble transition model $M_{\psi}$ and a sampling policy $\hat{\pi}_{\theta}$:

\begin{equation}
    \hat{g}_t = \bbE_{\hat{a}_{t+i}\sim\hat{\pi}_{\theta}(\hat{s}_{t+i}), \hat{s}_{t+i+1} \sim M_{\psi}}\left[\phi(\hat s_{t+K}) \right]~, 0\leq i \leq K-1~,
\end{equation}

where $\hat{a}_{t+i} \sim \hat{\pi}_{\theta}(\cdot|\hat{s}_{t+i},g_t)$ is the action sampled by $\hat{\pi}_{\theta}$ and $\hat{s}_{t+i+1} \sim M_{\psi}(\cdot|\hat s_{t+i}, \hat a_{t+i})$ is the state sampled from $M_{\psi}$, which is the probabilistic dynamics model ensemble mentioned above. Specifically, at each timestep, we randomly choose one of the dynamics models to get a 1-step rollout.

Besides, there are more details worth noting. First, FGI should take the first rollout step from $s_{t+1}$ when relabeling $s_t$ since it is the tuple $(s_t, a_t, g_t)$ that be used for training. Second, we take the rollout $H$ steps at maximum due to the model precision and leave the longer relabeling with the future strategy HER. We show the FGI algorithm step-by-step  in \alg{alg:FGI}.

We illustrate how FGI works in \fig{fig:fgi} compared with prior HER relabeling methods. 
FGI advances in several aspects: i) HER limits the relabeled goal in the sampled trajectory, yet FGI generates various goals determined by the dynamics model; 
ii) FGI regenerates goals using the current model and current policy even for the historical trajectories, while HER still depends on old ones, which may significantly differ from the current policy. 

\subsection{Universal Model-based Policy Optimization}

In this paper, we expand model-based policy optimization~\cite{janner2019trust} into the goal-based setting and propose Universal Model-based Policy Optimization (UMPO) for efficient goal-based policy learning. By extending the Q function $Q(s,a)$ to the universal Q function $Q(s,a,g)$ and applying DDPG~\cite{lillicrap2015continuous} as the policy optimization algorithm, the updating of the parameterized policy and Q function in UMPO can be written as follows:

\begin{equation}
\begin{aligned}
\label{pi_update}
    \nabla_{\theta} J\left(\pi_{\theta}\right) = 
    \mathop{\mathbb{E}}\limits_{\substack{(s,g)\sim \hat{\caD}\\a\sim\pi}}\left[\nabla_{a} Q_{\omega}(s, a, g) \nabla_{\theta} \pi_{\theta}(s,g) \right]
\end{aligned}
\end{equation}

\begin{equation}
\begin{aligned}
\label{Q_update}
    \nabla_{\omega}J(Q_{\omega}) =
    \mathop{\mathbb{E}}\limits_{\substack{(s,a,s',g) \sim \hat{\caD}\\a'\sim\pi}} &[ \nabla_{\omega}\left(Q_{\omega}(s,a,g) - r \right. \\ 
    & \left. - \gamma Q_{\omega'}(s', a', g)\right)^2] ~,
\end{aligned}
\end{equation}
where $Q_{\omega'}$ is the target Q function and $\hat{\caD}\triangleq\alpha\caD_{\text{env}}+(1-\alpha)\caD_{\text{model}}$ is a mixture of the real experience $\caD_{\text{env}}$ in the environment and generated short branched rollouts $\caD_{\text{model}}$ with the learned model. 

Unlike in classic RL environments, in GoRL setting, branched rollouts also need a behavioral goal $g$. Thus, how to choose appropriate goals when performing branched rollouts is a crucial problem. In UMPO, instead of taking the goal $g$ from the tuple $(s_t, a_t, s_{t+1}, g_t)$ directly, we randomly select a state $s_{t+n}$ from the same trajectory and replace $g_t$ with $\phi(s_{t+n})$ as in HER. It is believed that such a branched rollout strategy could improve the diversity of training samples and thus increase the performance, as evaluated in the experiment.

\section{Related Work}
\subsection{Goal-oriented Reinforcement Learning}
As mentioned above, GoRL algorithms roughly contain four steps: goal selection, rollout, relabeling, and optimization. An extensive set of previous works pay attention to the goal selection and the relabeling stage.

\textbf{Goal Selection.} The goal selection stage guides the agent to replace the raw desired goal with a currently appropriate one to achieve. For example, \citea{florensa2018automatic} proposed to utilize the generative adversarial network to generate the goals at appropriate difficulty. \citea{ren2019exploration} optimized the desired goals along with learning the policy, which selects them according to the result of the formulated Wasserstein Barycenter problem. \citea{hartikainen2019dynamical} chose maximum distance goals based on a learned distance function. And \citea{pitis2020maximum} sampled the goals which could maximize the entropy of the historical achieved goals.

\textbf{Relabeling.} The works for relabeling aims to increase sample efficiency and can be traced back to Hindsight Experience Replay (HER)~\cite{andrychowicz2017hindsight}, which shows that off-policy algorithms can reuse the data in the replay buffer by relabeling goals in the previous trajectory. As an improvement, \citea{fang2019curriculum} proposed CHER that adaptively selects the failed experiences for replay according to the proximity to true goals and the curiosity of exploration. Other works combine HER with other goals like generated goals by variational auto-encoder~\cite{nair2018visual} , past achieved goals, and actual desired goals~\cite{pitis2020maximum} given by the environments.

Besides, \citea{trott2019keeping} proposed a sibling rivalry (SR) mechanism to avoid getting to local optima and get a better exploration while sampling in the environment, which can not be classified into the former two categories.

\subsection{Model-Based Reinforcement Learning}
Our work adopts the dynamics model for foresight goal inference and further develops the usage of model-based techniques in GoRL. Model-Based Reinforcement Learning (MBRL) has been active in the community for years due to its high sample efficiency, which needs fewer interactions with the environment to derive a good policy. A common way for using the model is the Dyna-style algorithm~\cite{sutton1991dyna,feinberg2018model,buckman2018sample}, which improves the policy through sampling in the transition model; besides, Model Predictive Control (MPC)~\cite{camacho2013model,wang2019exploring,boney2020regularizing} does not maintain a policy but uses the model to look forward finite steps with several action sequences and chooses the first action with maximum cumulative rewards. 

Recently, \citea{luo2018algorithmic} built a lower bound of the expected reward and maximizes the lower bound jointly over the policy and the model. \citea{chua2018deep} adopted an ensemble of neural network models while applying MPC to sample good actions. Another important work comes from \citea{janner2019trust}, which analyzes the compounding errors in model rollouts and builds a lower bound of the value discrepancy between the model and the environment. The resulting algorithm named Model-Based Policy Optimization (MBPO) generates short rollouts from the real states and optimizes the policy with both the real and the generated samples.

Also, some methods absorbing the idea of MBRL are proposed in the field of GoRL. \citea{nair2020goal} built a goal-aware dynamics model to capture the dynamics in some multi-goal environments with only image inputs. \citea{pitis2020counterfactual} learned local casual models to generate counterfactual experiences, which significantly improves the performances of RL agents in locally factored tasks.

\section{Experiments}

\begin{figure*}[!htbp]
\centering
\subfigure[Episode 200]{
\begin{minipage}[b]{0.2\linewidth}
\label{fig:world-iter-200}
\includegraphics[width=1\linewidth,trim={0 0 0 0cm},clip]{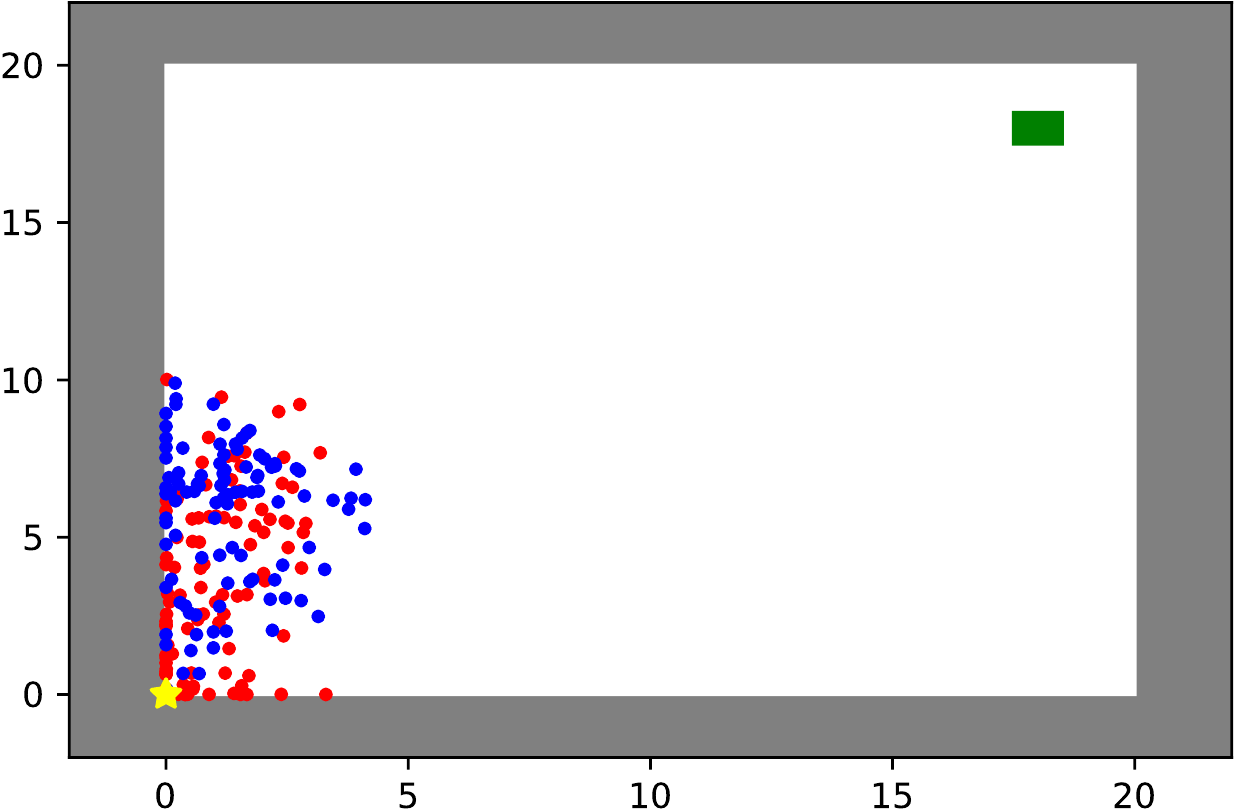}
\end{minipage}
}
\hspace{-10pt}
\subfigure[Episode 500]{
\begin{minipage}[b]{0.2\linewidth}
\label{fig:world-iter-500}
\includegraphics[width=1\linewidth,trim={0 0 0 0cm},clip]{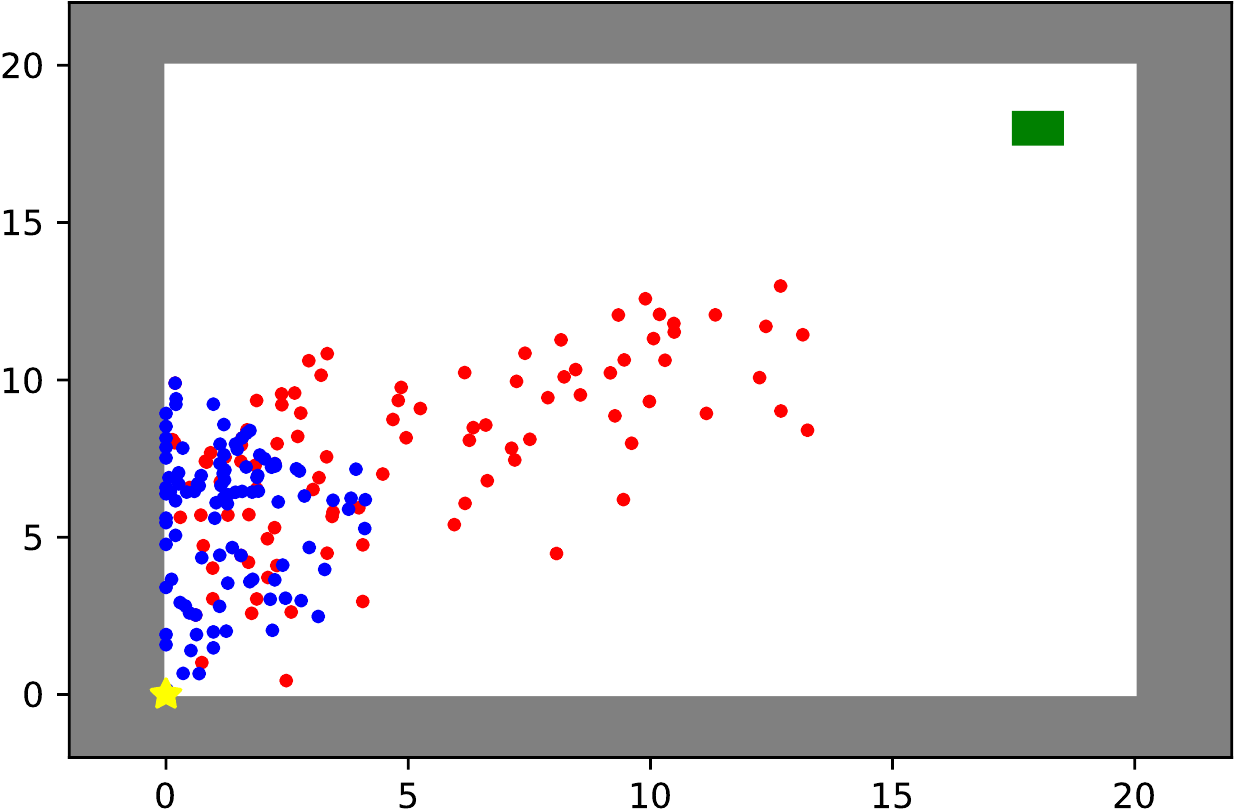}
\end{minipage}
}
\hspace{-10pt}
\subfigure[Episode 1000]{
\begin{minipage}[b]{0.2\linewidth}
\label{fig:world7iter-1000}
\includegraphics[width=1\linewidth,trim={0 0 0 0cm},clip]{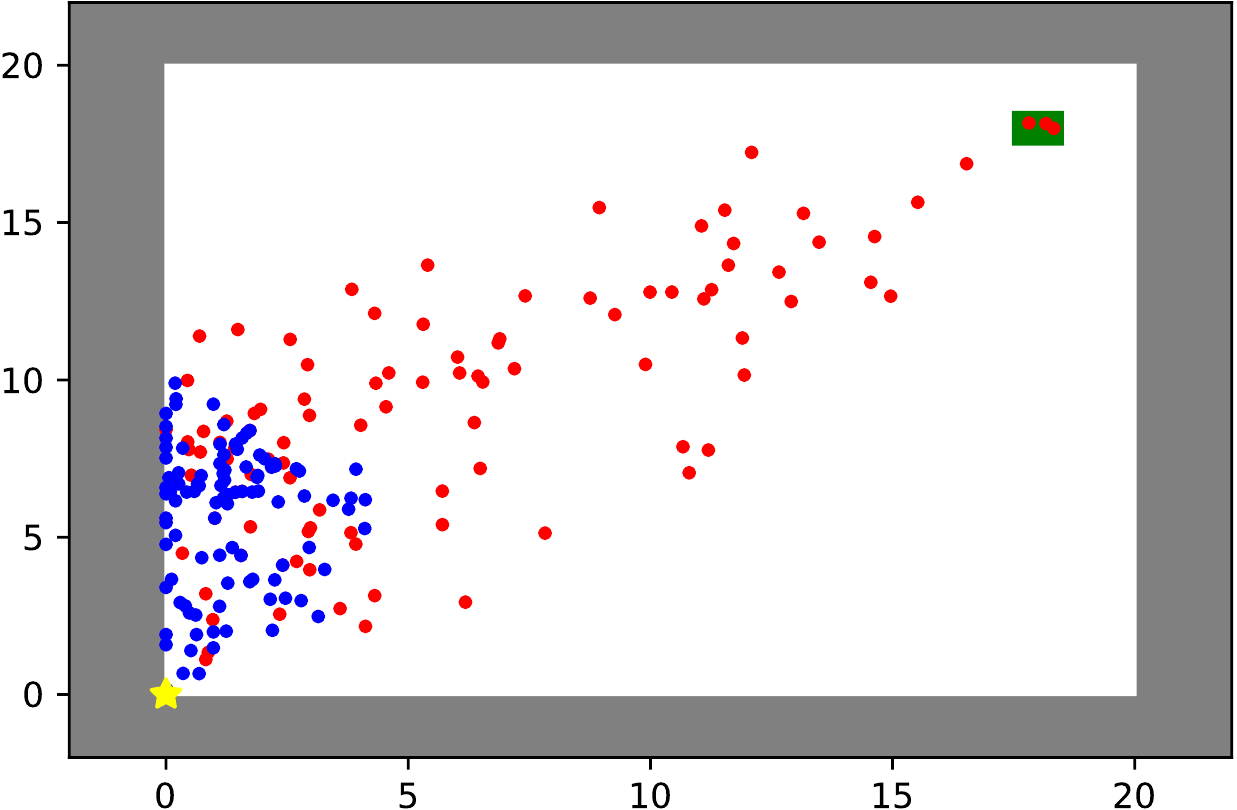}
\end{minipage}
}
\hspace{-10pt}
\subfigure[Episode 1300]{
\begin{minipage}[b]{0.2\linewidth}
\label{fig:world-iter-1300}
\includegraphics[width=1\linewidth,trim={0 0 0 0cm},clip]{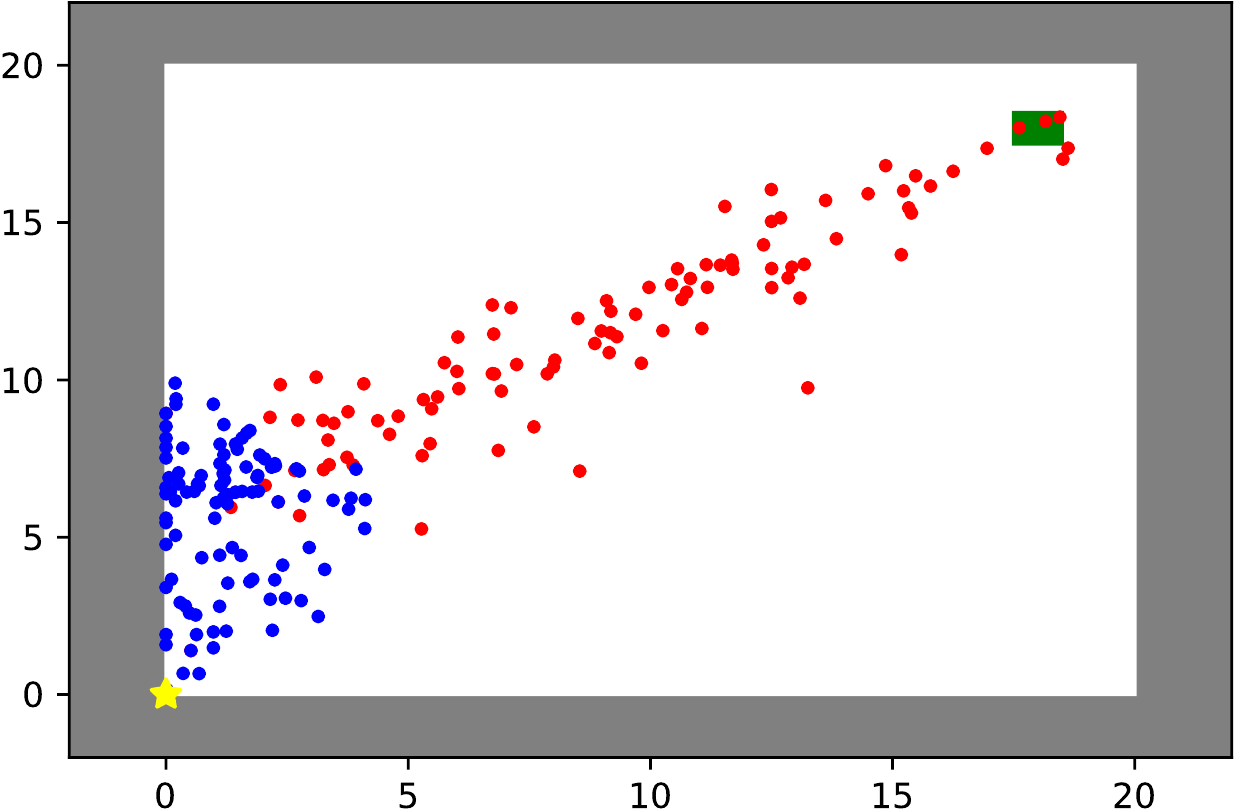}
\end{minipage}
}
\hspace{-5pt}
\begin{minipage}[b]{0.15\linewidth}
\label{fig:world-legend}
\includegraphics[width=1\linewidth]{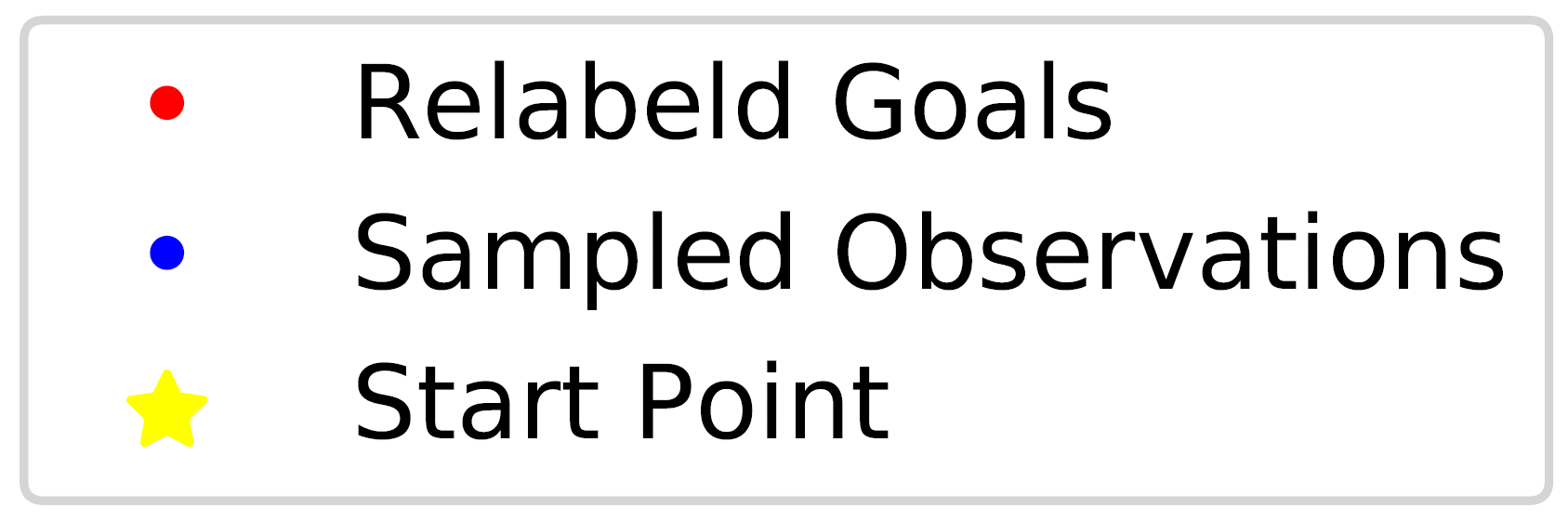}
\end{minipage}
\caption{Relabeled goal points (red) and on the same trajectories points (blue) on 2D-World during training. Agents tries to learn to go to the goal (green square) from one start point (yellow star). Note that the trajectories are sampled by an initialized policy and are saved in the replay buffer for training. It is obvious that FGI relabel different goals according to the current policy even for the same trajectories, which provides much more information for the learning policy. However, HER is only constrained in the sampled trajectories (blue) at every training stage.}
\label{fig:world-goals}
\end{figure*}

In this section, we conduct several experiments from different dimensions to answer the following questions:

\begin{enumerate}
    \item Does our FGI relabeling trick work as expected and is more advanced than HER?
    \item Can MapGo show advantage over previous methods on complicated goal-oriented tasks?
    \item Which component of MapGo is the most effective?
\end{enumerate}

To answer the first question, we demonstrate the effectiveness of FGI in a simple 2D locomotion task.
We show the relabeled goals by FGI during different training procedures and the learning curve compared to HER. To answer the second question, we evaluate the performance of the proposed MapGo architecture on four complicated goal-oriented benchmarks. To answer the third question, we conduct an ablation study where we compare the functionality of FGI and UMPO separately. We also make further analysis on model rollout strategies and the limitations of MapGo.

\subsection{Relabeling Goals using FGI}
We first show that our FGI relabeling trick offers fresher information for policy training than HER does. We conduct the comparison experiments on 2D-World, a simple 2D navigation task where the state space is the position of the agent $(x,y)\in[0,20]^2$, and the action space is the movement $(\delta_x,\delta_y)\in[-1, 1]^2$. The initial state $s_0=(0, 0)$ is fixed in all episodes. This task aims to reach a target that is randomly sampled in $[18.5,19.5]^2$, and the agent receives a nonnegative reward only when the agent gets close enough to the target. 
We choose the maximum rollout length $H$ as 20 and do not use extra relabeling methods from HER. 

\begin{figure}[htbp]
    \centering
    \includegraphics[width=0.9\columnwidth]{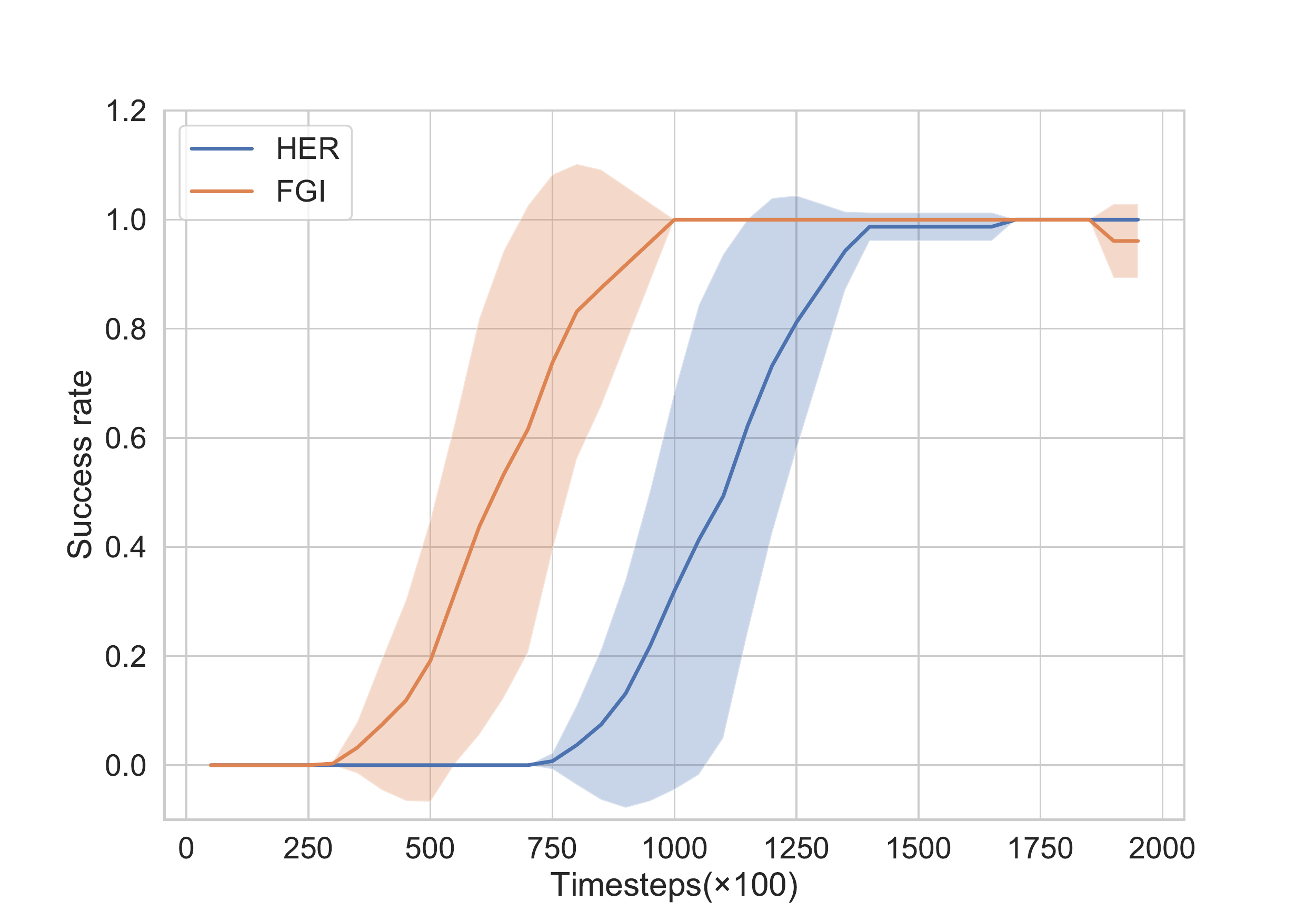}
    \caption{Training curves on 2D-World, where the solid line and the shade represent the mean and the standard deviation of the success rate separately, which is computed over 5 different random seeds. This shows the advantage of FGI which helps agent learn faster compared with HER. }
    \label{fig:world-curve}
\end{figure}

We illustrate the relabeled goals by FGI at different training stages in \fig{fig:world-goals}, where the blue points represent trajectories sampled by an initialized policy for off-policy training and the reds are the relabeled goals. As the training goes on, the relabeled goals are moving closer to the goal area, even for the same trajectories. It indicates that FGI relabels goals adapting the current policy and can provide much more useful information. On the other hand, HER can only relabel goals that coincide with the data in previous trajectories. As shown in \fig{fig:world-goals}, for a previous trajectory (blue points), HER still relabels goals from them without any additional information as the policy changes. The training curves are shown in \fig{fig:world-curve}, verifying the advantage and efficacy of FGI.

\begin{figure*}[hbtp]
\centering
\subfigure[Reacher]{
\begin{minipage}[b]{0.235\linewidth}
\includegraphics[width=1\linewidth]{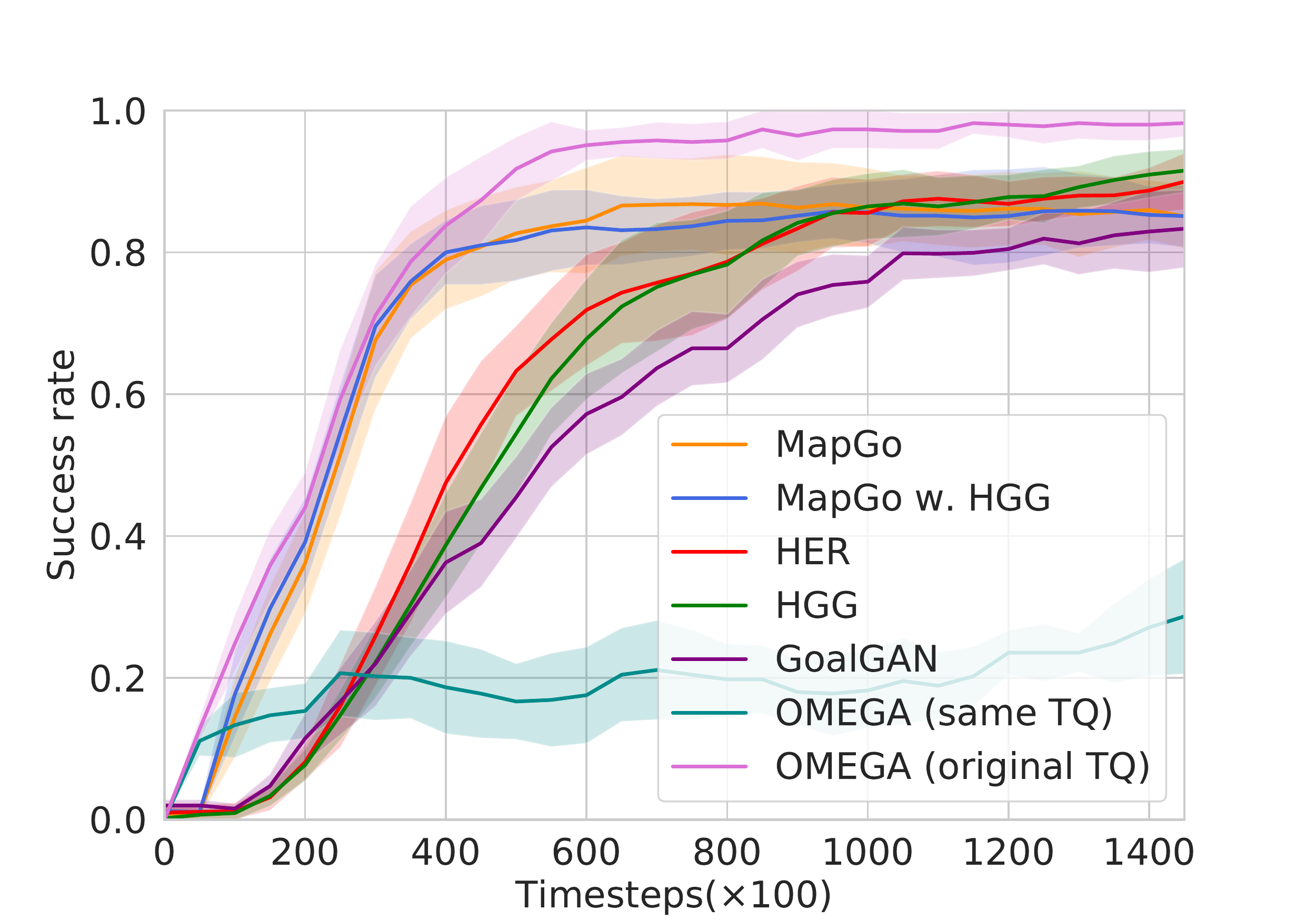}
\end{minipage}
}
\subfigure[HalfCheetah]{
\begin{minipage}[b]{0.235\linewidth}
\includegraphics[width=1\linewidth]{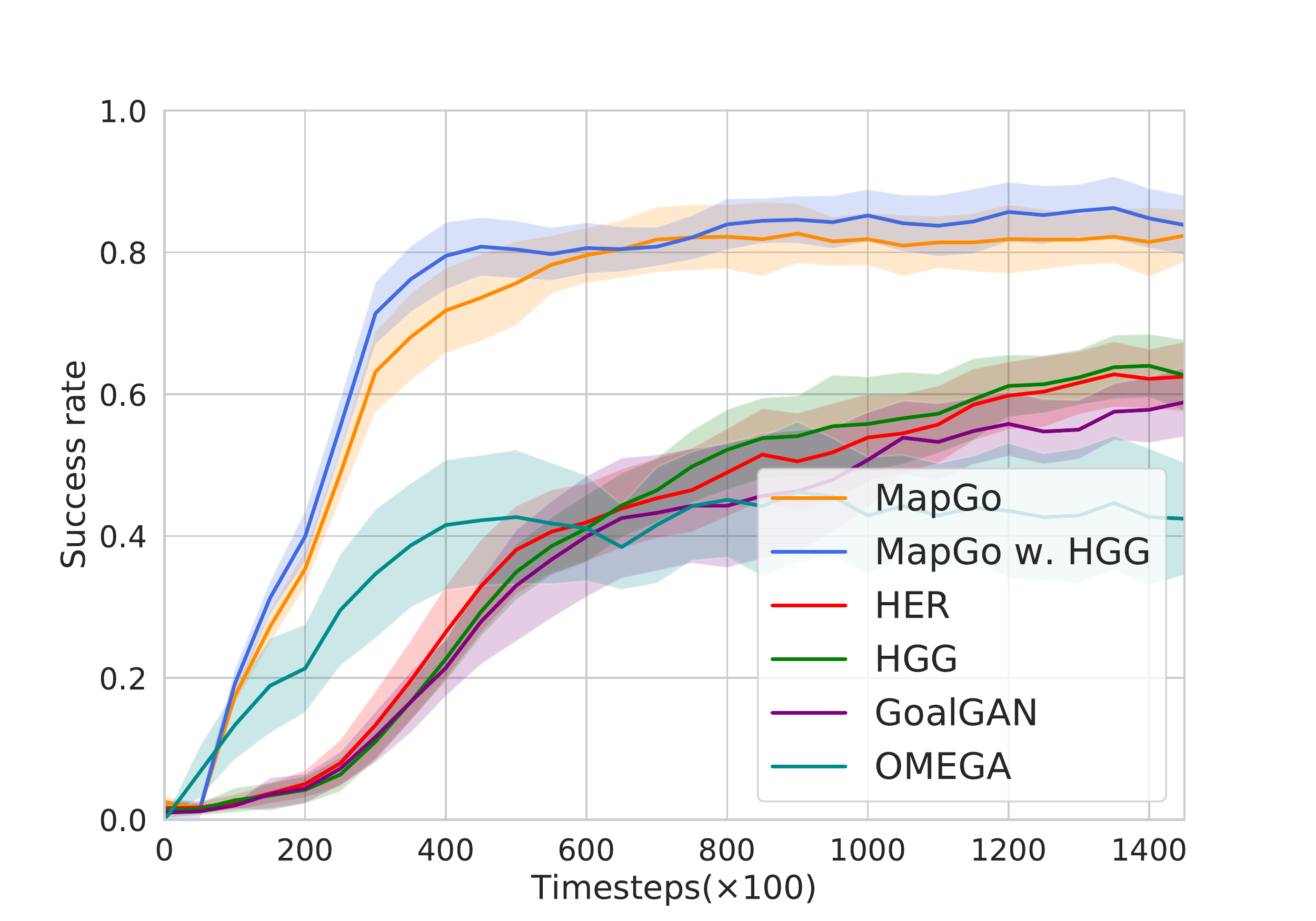}
\end{minipage}
}
\subfigure[Diverse Ant Locomotion]{
\begin{minipage}[b]{0.235\linewidth}
\includegraphics[width=1\linewidth]{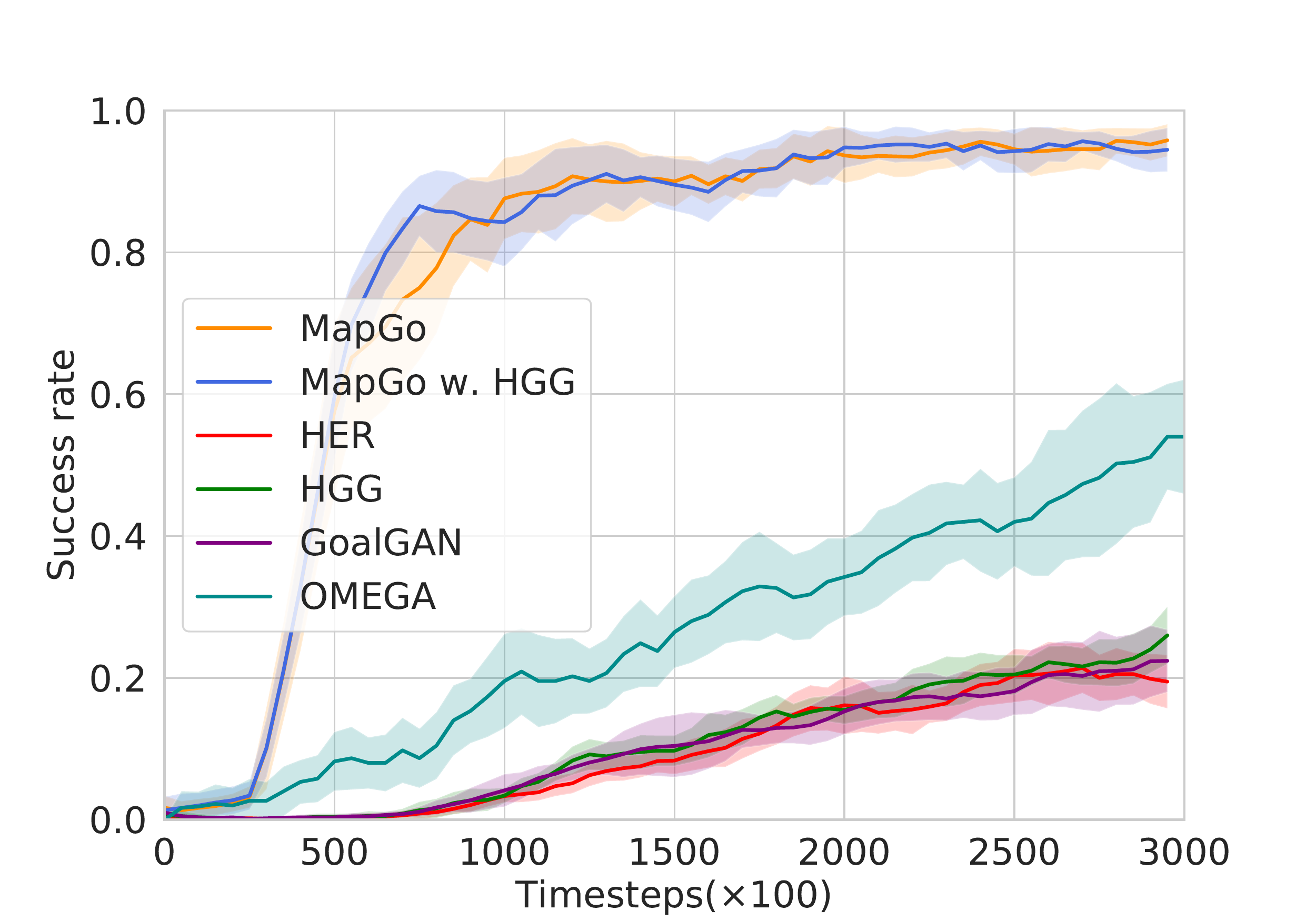}
\end{minipage}
}
\subfigure[Fixed Ant Locomotion]{
\begin{minipage}[b]{0.235\linewidth}
\includegraphics[width=1\linewidth]{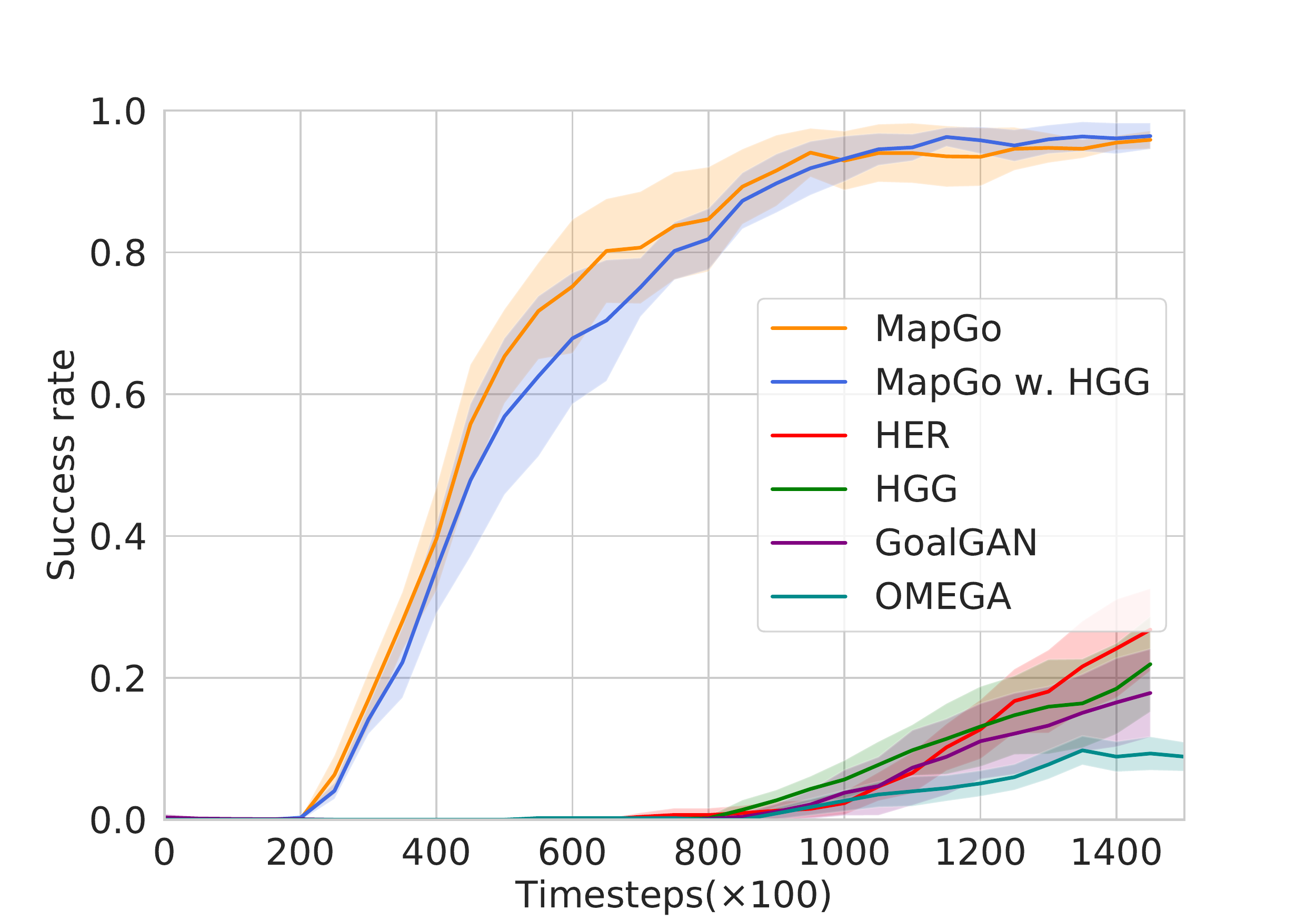}
\end{minipage}
}
\caption{Learning curves of MapGo (ours) compared with baselines on three goal-oriented tasks. MapGo w. HGG denotes MapGo using HGG for behavioral goal selection. The solid line indicates the mean and the shaded area is the standard error over 5 different random seeds. Each trial is evaluated every 5000 environment steps, where each evaluation reports the average return over 100 episodes.}
\label{fig:benchmark_curve}
\end{figure*}

\subsection{Complicated Goal-oriented Tasks}
In this section, we aim to show the advantage of MapGo compared with previous methods on four more challenging goal-oriented tasks: Reacher, HalfCheetah, Fixed Ant Locomotion, and Diverse Ant Locomotion. In Reacher, we control a 2D robot to reach a randomly located target~\cite{charlesworth2020plangan}. 
Fixed Ant Locomotion and Diverse Ant Locomotion are similar to the environment in \cite{florensa2018automatic}, where the agent is asked to move to a target position.

The main difference between them is whether the target is around a fixed far-away area or in the whole state space. It is worth noting that a fixed area does not mean that the environment is easier. A diverse distribution in another way provides the agent with the chance to learn from similar goals, while the fixed point desired goal makes the rewards even more sparse. HalfCheetah resembles the environment in~\cite{finn2017model} and requires the agent to run while keeping a targeted speed until the end of the episodes. 

In detail, we compare MapGo with HER, Hindsight Goal Generation (HGG)~\cite{hartikainen2019dynamical}, GoalGAN~\cite{florensa2018automatic} and OMEGA~\cite{pitis2020maximum}. We also test the performance of MapGo with HGG selecting the behavioral goal at each initial state (denoted as MapGo w. HGG). In Reacher, the rollout maximum length is $50$ and in the others, we set it as $30$. We utilize DDPG~\cite{lillicrap2015continuous} as the learning algorithm for all the methods.

We plot the learning curve in \fig{fig:benchmark_curve}, where MapGo outperforms all the other baselines and achieves higher sample efficiency in the most environments. Specifically, in Diverse Ant Locomotion, our algorithm takes about $2.0 \times 10^5$ timesteps to get an averaged success rate of 0.95, while the best baselines OMEGA is still at about 0.4 at the same time. Moreover, in HalfCheetah, MapGo learns much faster and only takes about 40 thousand steps to get a good performance. The only exception is that in Reacher, OMEGA performs better than MapGo because of a 50 times training frequency. For fairness we test OMEGA with different training frequencies (denoted as \textit{original TQ} and \textit{same TQ}), and show that MapGo actually still outperforms it with the same training frequency. We also find that the goal selection technique makes limited improvements on MapGo. One possible reason is that the model rollouts could provide enough exploration ability as HGG does.

\begin{figure}[htbp]
    \centering
    \includegraphics[width=0.85\columnwidth,trim={0 0 0 1.95cm},clip]{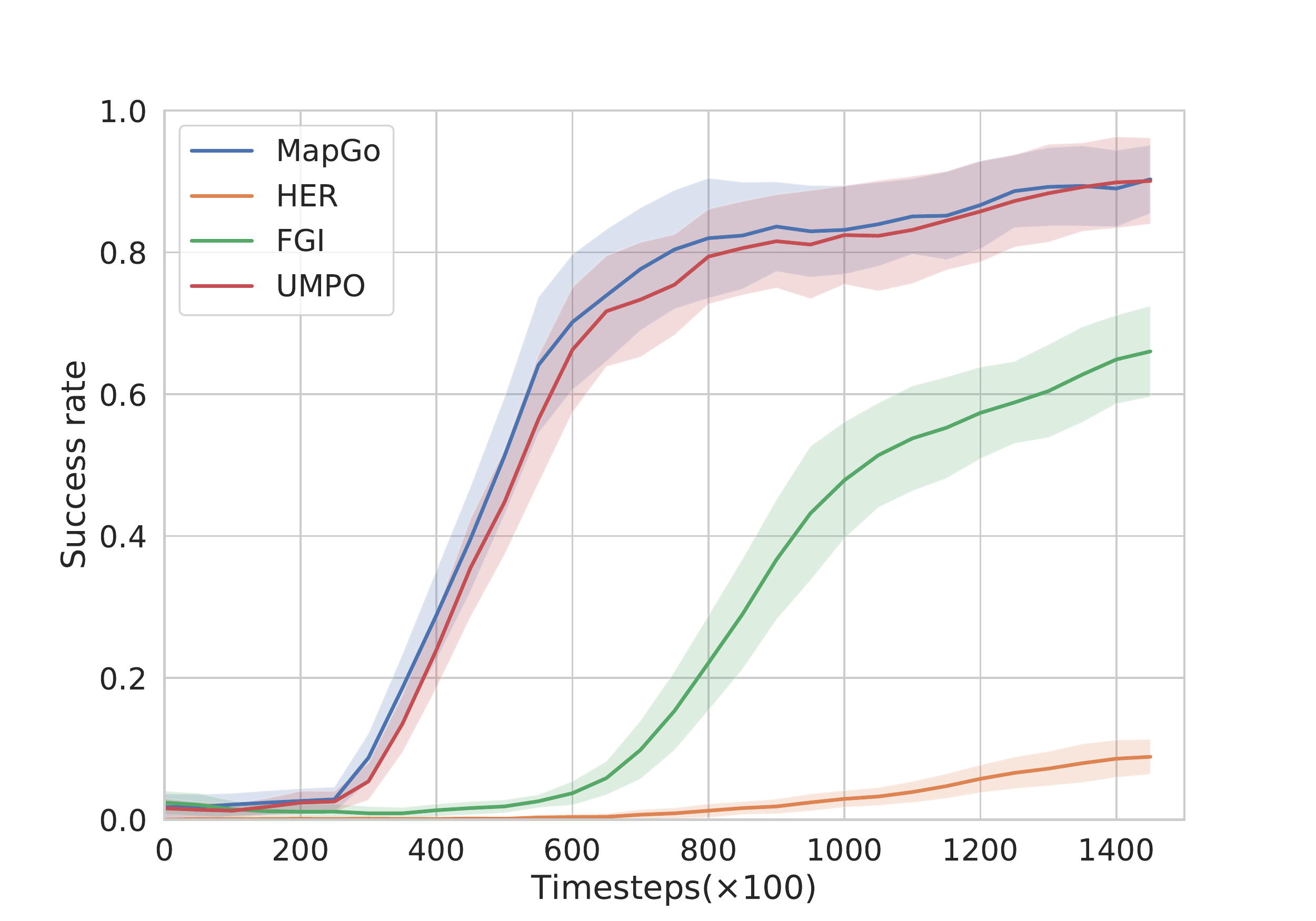}
    \caption{Ablation experiments on Diverse Ant Locomotion over 5 different random seeds. We analyse which component of MapGo works most by running UMPO and FGI respectively. It is obvious that UMPO contributes much more but by combining UMPO with FGI, MapGo achieves the best sample efficiency.}
    \label{fig:ablation}
\end{figure}

\subsection{Ablation Study}
\label{sec:ablation1}
To figure out how FGI and UMPO contribute to the overall performance of MapGo, we perform an ablation study on Diverse Ant Locomotion. Specifically, we compare the performance of the whole MapGo as conducted in the last section with running UMPO and FGI separately. Note that we do not use any relabel or goal selection techniques for UMPO.

As shown in \fig{fig:ablation}, it is easy to conclude that UMPO is more effective, which can achieve close sample efficiency as MapGo. However, with combining FGI and UMPO, MapGo still improves and achieves the best sample efficiency. Besides, compared to HER, FGI still outperforms it substantially, which proves its effectiveness.

\subsection{Different Branched Rollout Strategies}
\label{sec:ablation2}
\begin{figure}[htbp]
    \centering
    \includegraphics[width=0.85\columnwidth,trim={0 0 0 1.95cm},clip]{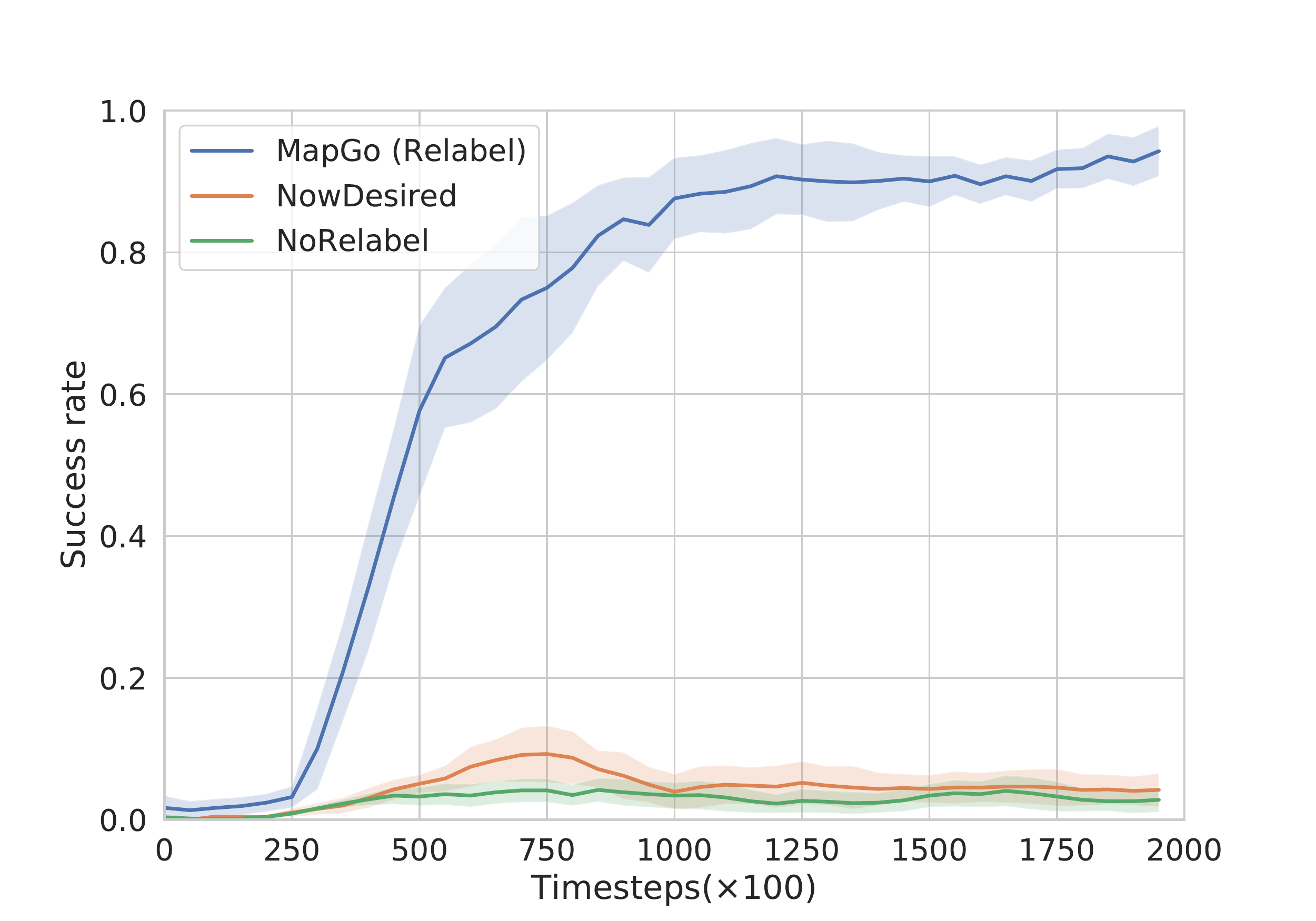}
    \caption{Ablation experiments on Diverse Ant Locomotion over 3 different random seeds. We analyse the model rollout strategy of MapGo frameworks and we compare MapGo with \textit{NoRelabel} and \textit{NowDesired} strategy.
    }
    \label{fig:ablation2}
\end{figure}

In MapGo, to encourage the exploration and increase the diversity in model rollouts, we utilize hindsight relabeling to set a behavioral goal before sampling with the model. In this section, we test it with other different rollout strategies on Diverse Ant Locomotion: 1) \textit{NoRelabel}, taking the original desired goal from the environment for rollouts; 2) \textit{NowDesired}, choosing the behavior goals using a current HGG goal selector. As shown in \fig{fig:ablation2}, the relabel strategy used in MapGo when performing branched rollouts is much more effective than the others. This implies that the goals should be diverse to achieve better performance. 

\subsection{Limitations of MapGo}

Although MapGo provides visible improvements in sample efficiency on complicated goal-oriented tasks, we surprisingly find it performs poorly for some robotic tasks. To understand such a situation, we compare the loss of the learned models on Ant locomotion and FetchPush and draw the learning curves in Appendix, \fig{fig:model_loss}. 
It shows that the current way of learning a dynamics model is unhelpful for such a robotic task, and the loss diverges. 

This can be attributed to the fact that the transition function in such environments is somewhere discontinuous. 

For example, small movements of robots could separate the arm and the object, thus significantly affecting the transition function. 
Since MapGo relies heavily on the learned model, which serves to both FGI and UMPO, this can lead to poor performances.
Similar observations can be found in \citea{pitis2020counterfactual}, where the authors try to build a mask to represent the adjacency matrix of the sparsest local causal graph by dynamics model. 
However, in robotics tasks like \textit{FetchPush}, all the masks are generated by domain knowledge rather than a dynamics model. It implies that the dynamics model limits the application of MapGo only if we can improve the accuracy of the learned model in these environments.

\section{Conclusion}

We tackle the problem of GoRL and aim to solve the sample efficiency and the diversity of relabled goals. Specifically, we propose MapGo, which is integrated by a novel relabeling strategy named FGI that equips the agent with the foresight ability by looking into the future with the current policy, and UMPO that applies branch rollout to increase the sample efficiency. Qualitative and quantitative experiments show the effectiveness of MapGo and evaluate each component comprehensively. In the future, we will further study cases of discontinuous transitions when using dynamics models. 

\section*{Acknowledgement}
This work is supported by the ``New Generation of AI 2030'' Major Project (2018AAA0100900), National Natural Science Foundation of China (62076161, 61632017) and the Tencent Rhino-Bird Research Project.

\bibliographystyle{named}
\bibliography{ijcai21}

\clearpage

\onecolumn
\newpage
\appendix

\setcounter{lemma}{0}
\setcounter{theorem}{0}
\section{Appendix}

\subsection{Universal Model-based Policy Optimization}
\label{ap:alg-umpo}
\begin{algorithm}[h]
\caption{Universal Model-based Policy Optimization}
\label{alg:umpo}
\begin{algorithmic}
\REQUIRE Policy $\pi_\theta$, Q function $Q_\omega$, dataset from env $\caD_{\text{env}}$, training batch from env $\caD_{\text{real}}$ and dynamics model $M_\psi$.
\FOR{iteration=1 to N}
    \STATE $\caD_{\text{fake}} \leftarrow \varnothing$;
    \FOR{i=1 to K}
        \STATE Sample trajectories $\tau$ from $\caD_{\text{env}}$;
        \STATE Randomly select a state $s_t$ and a new desired goal $g$ from $\tau$ with HER;
        \STATE Perform $k$-step rollout starting from $s_t$ with $g$ using policy $\pi_{\theta}$ and dynamics model $M_\psi$; add to $\caD_{\text{fake}}$;
    \ENDFOR
    \STATE Update policy parameters $\theta$ and Q function parameters $\omega$ with dataset $\caD_{\text{real}} \cup \caD_{\text{fake}}$ according to equation \ref{pi_update} and \ref{Q_update};
\ENDFOR
\end{algorithmic}
\end{algorithm}

\subsection{Details of Environments}

\subsubsection{2D-World}
As described above, in this environment, we control a point in 2D space from the initial state $s_0=(0, 0)$ to the goal square. The target is sampled from $[18.5, 19.5]$. The action space is $[-1, 1]^2$ and the maximum length of an episode is $100$. An episode is judged as succeeded if and only if the point stays close enough to the target in the final step. Here we set the distance threshold as 0.15. The dimension of the action space, state space and goal space are all 2.

\subsubsection{Reacher}
In this environment, we control a 2D robot to reach to a randomly located target. The code we used is from the released code of PlanGAN\cite{charlesworth2020plangan}. The dimension of action space is 2, the dimension of state space is 11 and the dimension of goal space is 2. The maximum length of an episode is $100$ and the distance threshold of success is set to $0.02$. An episode is judged as succeeded if and only if the agent receives nonnegative reward at the final timestep.

\subsubsection{HalfCheetah}
This environment is based on \cite{todorov2012mujoco}, where we control a robot to keep the speed in the target range. The dimension of action space is 6, the dimension of state space is 17 and the dimension of goal space is 1. The maximum length of an episode is $100$ and the distance threshold of success is set to $0.1$. An episode is judged as succeeded if and only if the agent receives nonnegative reward at the final timestep.

\subsubsection{Diverse Ant Locomotion and Fixed Ant Locomotion}
These two environments are similar and both based on \cite{todorov2012mujoco}. We control an ant robot to get to the target place. The dimension of action space is 8, the dimension of state space is 29 and the dimension of goal space is 2. The only difference between these two environments is that target in Diverse Ant Locomotion is chosen from the box space from $[-2, -2]$ to $[2,2]$, while the target in Fixed Ant Locomotion is fixed to $[2,2]$. An episode is judged as succeeded if and only if the agent receives nonnegative reward at some timestep in the episode.

\subsection{Details of Implementation}

The code of dynamics model is based on the realization of \cite{janner2019trust} and we modified it slightly to fix the bug that using an increasing number of video memory. The memory pool we used is actually two queues which stores the tuples $(s,a,s',g)$. The one stores the trajectories collected from the environment while the other stores the generated trajectories by model rollouts. When training, we samples different numbers of tuples from the two memory buffers according to the batch size and the mixture factor $\alpha$.

In our experiments, both HGG and OMEGA are based on their official implementation\footnote{HGG: https://github.com/Stilwell-Git/Hindsight-Goal-Generation}\footnote{OMEGA: https://github.com/spitis/mrl}. HER is based on the code of HGG, where we removed the goal selector in HGG. As for GoalGAN, we borrowed the goal generator from its original implementation \footnote{GoalGAN: https://sites.google.com/view/goalgeneration4rl} and combined it with HER to get our implementation of baseline GoalGAN.

\subsection{Hyperparameters}
\begin{table}[htbp]
  \centering
  \caption{Hyperparameters}
    \begin{tabular}{|l|c|c|c|c|c|}
    \hline
    Environments & \multicolumn{1}{l|}{2D-World} & \multicolumn{1}{l|}{Reacher} & \multicolumn{1}{l|}{HalfCheetah} & \multicolumn{1}{l|}{Diverse Ant Locomotion} & \multicolumn{1}{l|}{Fixed Ant Locomotion} \\
    \hline
    Rollout maximum length $H$& \multicolumn{1}{c|}{20}& \multicolumn{1}{c|}{50} & \multicolumn{3}{c|}{30} \\
    \hline
    Episode length $h$& \multicolumn{5}{c|}{100} \\
    \hline
    Optimizer & \multicolumn{5}{c|}{AdamOptimizer} \\
    \hline
    Discount factor $\gamma$& \multicolumn{5}{c|}{0.98} \\
    \hline
    Memory buffer size & \multicolumn{5}{c|}{100000} \\
    \hline
    Batch size & \multicolumn{5}{c|}{256} \\
    \hline
    Training batch mixture factor $\alpha$ & \multicolumn{5}{c|}{0.05} \\
    \hline
    Q learning rate & \multicolumn{1}{c|}{1e-4} & \multicolumn{4}{c|}{1e-3} \\
    \hline
    $\pi$ learning rate & \multicolumn{1}{c|}{5e-4} & \multicolumn{4}{c|}{1e-3} \\
    \hline
    Transition model learning rate & \multicolumn{5}{c|}{1e-3} \\
    \hline
    Model training validation set factor & \multicolumn{5}{c|}{0.2} \\
    \hline
    Boopstrapping model number & \multicolumn{5}{c|}{6} \\
    \hline
    Elite model number & \multicolumn{5}{c|}{3} \\
    \hline
    \end{tabular}%
  \label{tab:addlabel}%
\end{table}%

\subsection{Additional Experiment Results}

\begin{figure}[!htbp]
    \centering
    \includegraphics[width=0.6\columnwidth]{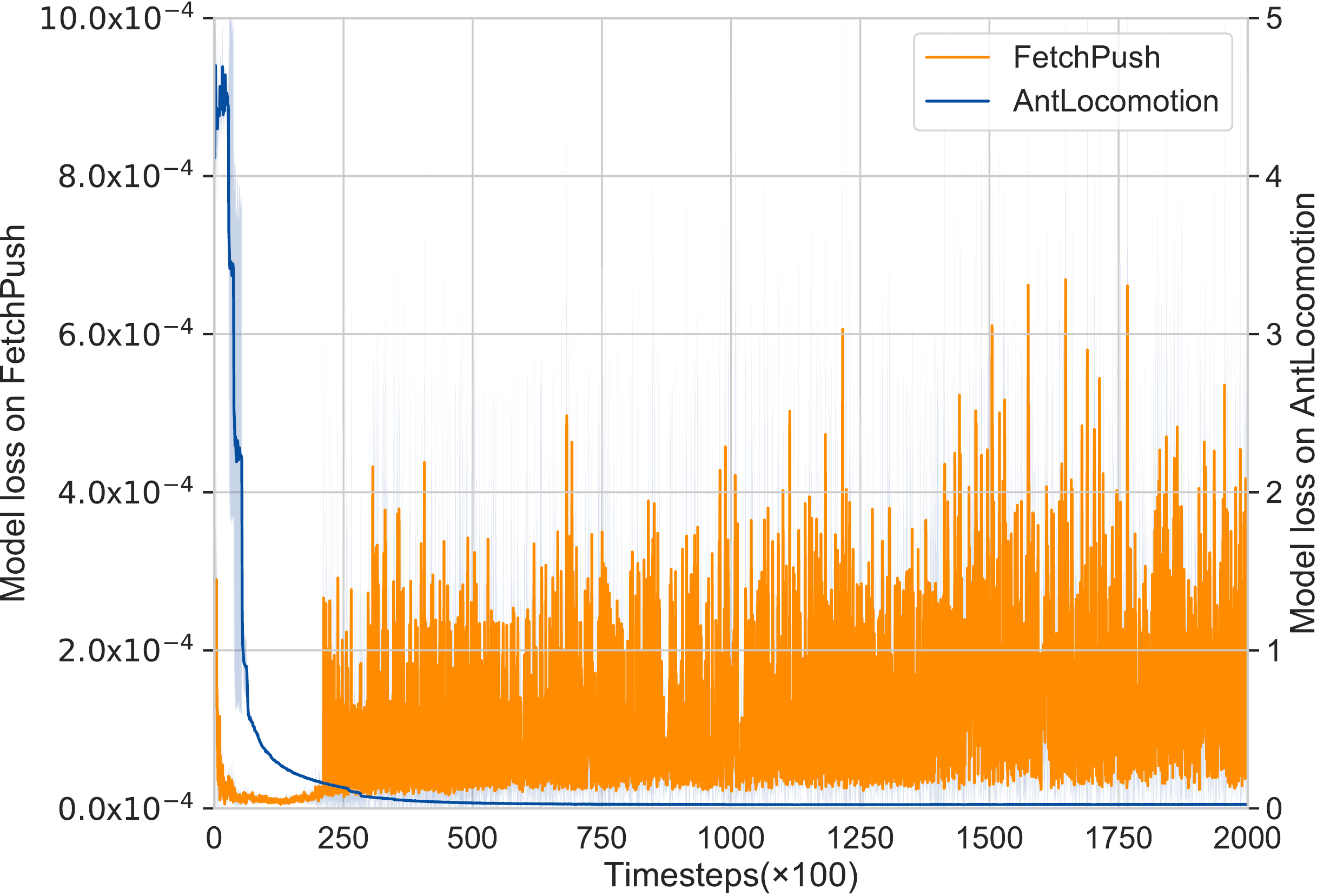}
    \vspace{-5pt}
    \caption{The loss of the learned model on two different environments. It can be easily observed that the model is learned well on the Ant Locomotion but for Fetch Push, the loss of the learned model diverges.}
    \label{fig:model_loss}
    \vspace{-5pt}
\end{figure}

\end{document}